\documentclass[letterpaper, 10pt, conference]{ieeeconf}
\IEEEoverridecommandlockouts    
\usepackage{bm}
\usepackage{graphicx}
\usepackage{float}
\usepackage{subfig}
\usepackage{amsmath} 
\usepackage{amssymb}  
\usepackage{amsfonts}

\usepackage{amsthm}

\usepackage{booktabs}                                   
\usepackage{multirow}                                   
\usepackage{makecell}                                   
\usepackage{tablefootnote}                              
\usepackage[symbol]{footmisc}                           
\usepackage{amsmath,amssymb}                            
\usepackage{xcolor}                                     
\let\labelindent\relax 
\usepackage{enumitem}                                   
\usepackage{cite}

\newcommand{\etal}{\textit{et al.~}}

\title{\LARGE \bf
CMG-Net: An End-to-End Contact-Based\\Multi-Finger Dexterous Grasping Network
}

\author{
  Mingze Wei$^{1,*}$, Yaomin Huang$^{2,*}$, Zhiyuan Xu$^{3}$, Ning Liu$^{3}$, Zhengping Che$^{3}$, \\Xinyu Zhang$^{1}$, Chaomin Shen$^{2}$, Feifei Feng$^{3}$, Chun Shan$^{4}$, and Jian Tang$^{3}$
  \thanks{
    $^{1}$School of Software Engineering, East China Normal University, China.
    {51205902058@stu.ecnu.edu.cn, xyzhang@sei.ecnu.edu.cn}
  }%
  \thanks{
    $^{2}$School of Computer Science, East China Normal University, China.
    {51205901049@stu.ecnu.edu.cn, cmshen@cs.ecnu.edu.cn}
  }%
  \thanks{
    $^{3}$Midea Group, China.
    {
    xuzy70@midea.com, liuning22@midea.com, chezp@midea.com, feifei.feng@midea.com, tangjian22@midea.com
    }
  }%
  \thanks{
    $^{4}$School of Electronics and Information, Guangdong Polytechnic Normal University, China.
    {shanchun@gpnu.edu.cn}
  }%
  \thanks{
    $^{*}$The first two authors contributed equally. This work was done when Mingze Wei and Yaomin Huang took internships at Midea Group.
  }
  \thanks{
    Corresponding authors: Xinyu Zhang and Jian Tang.
  }%
}

\begin{document}

\maketitle
\thispagestyle{empty}
\pagestyle{empty}

\begin{abstract}
In this paper, we propose a novel representation for grasping using contacts between multi-finger robotic hands and objects to be manipulated. This representation significantly reduces the prediction dimensions and accelerates the learning process.
We present an effective end-to-end network, CMG-Net,
for grasping unknown objects in a cluttered environment by efficiently predicting multi-finger grasp poses and hand configurations from a single-shot point cloud.
Moreover, we create a synthetic grasp dataset that consists of five thousand cluttered scenes, 80 object categories, and 20 million annotations. We perform a comprehensive empirical study and demonstrate the effectiveness of our grasping representation and \mbox{CMG-Net}. Our work significantly outperforms the state-of-the-art for three-finger robotic hands.
We also demonstrate that the model trained using synthetic data performs very well for real robots.
\end{abstract}

\section{Introduction}\label{sec:intro}
Grasping unknown objects from a cluttered environment is a fundamental problem for autonomous robotic manipulation, arising in a wide range of applications such as industrial automation (e.g., bin-picking, quality inspection, and warehouse automation), commercial places (e.g., book picking or placing in a library and healthcare) and household (e.g., folding laundry, playing billiards, and fetching a beer in a refrigerator). Despite the exciting progress in pose prediction and object manipulation for parallel-jaw grippers~\cite{DBLP:conf/cvpr/FangWGL20,jiang2011efficient,DBLP:journals/ijrr/PasGSP17,DBLP:conf/iccv/MousavianEF19,DBLP:conf/icra/MuraliMEPF20,DBLP:conf/icra/SundermeyerMTF21}, robotic grasping for multi-finger hands with high DoFs remains challenging. Parallel-jaw grippers have relatively low complexity in manipulation due to their one DoF structure. However, their structures also limit their deployment in more complicated scenarios with relatively big, irregular, or spherical objects.

Unlike parallel-jaw grippers, multi-finger dexterous hands significantly improve adaptability for object shapes due to higher DoFs and flexibility.
However, the difficulties in finding valid hand configurations and grasp poses for multi-finger robotic hands are greatly increased due to high-dimensional search space and discontinuous grasp space~\cite{ciocarlie2007dexterous,DBLP:conf/rss/LiuP0GM20,DBLP:conf/iros/HangSK14}.

To handle these issues, previous work for multi-finger grasping can be divided into two categories.
\emph{Traditional analytical methods}~\cite{ciocarlie2007dexterous,DBLP:conf/icra/GoldfederALP07,DBLP:conf/iros/HangSK14,DBLP:conf/icra/MillerKCA03,DBLP:conf/icra/PelossofMAJ04} explore the potential grasping space through stochastic search and sampling.
These algorithms are often computationally expensive, typically requiring tens or even hundreds of iterations per object. Moreover, they heavily rely on precise object representation, thus not applicable to unknown objects in a cluttered environment.
\emph{Data-driven approaches} have attracted attention in recent years.
Lu~\etal proposed a grasp planner and grasp evaluator based on an efficient deep neural network~\cite{DBLP:conf/iros/BrahmbhattHHF19}. This work still assumes the object models are known.
Lundell~\etal proposed a coarse-to-fine model to predict grasps~\cite{DBLP:conf/icra/LundellCLVWRMK21}. This work targets a single object instead of many objects in a cluttered environment.
Li \etal presented an end-to-end network~\cite{DBLP:conf/icra/LiWL0LZ22} that can predict grasp poses and configurations. It simply discretizes and classifies multi-finger hand configurations manually.

In this paper, we propose a novel approach to grasp unknown objects in a cluttered environment. Our contribution includes:
\begin{itemize}
\item A new grasp representation that projects 10-DoFs grasps to only 6-DoFs based on contact points. This significantly reduces the potential grasping search space, facilitates the process of learning and improves the grasping quality.
\item Based on this novel representation, we propose an end-to-end deep neural network, CMG-Net, which outputs multi-finger hand configurations and grasp poses for an input single-shot viewpoint in a cluttered scene.
\item To demonstrate the benefits using CMG-Net, we build a synthetic grasping dataset, consisting of 5000 cluttered scenes with 80 object classes and 20 million grasp annotations (i.e., hand poses and hand configurations).
\item Our experimental results show that CMG-Net outperforms the state-of-the-art in terms of both grasping success rate and grasping quality. Moreover, we also demonstrate that CMG-Net works very well for robotic grasping in a real-world environment.
\end{itemize}

\section{Related Work}\label{sec:related}

Grasping is a fundamental problem for robotic manipulation and has been extensively studied. Most work focuses on parallel-jaw grippers \cite{DBLP:conf/cvpr/FangWGL20,jiang2011efficient,DBLP:conf/iccv/MousavianEF19,DBLP:conf/icra/MuraliMEPF20,DBLP:conf/icra/SundermeyerMTF21}  due to their simplicity, low DoFs, and computational efficiency. However, parallel-jaw grippers are less efficient and less reliable for manipulating arbitrary-shaped objects. To achieve user-friendly interaction, multi-finger robotic hands and dexterous grasping remain a hot research topic in the field of robotic manipulation~\cite{rimon2019mechanics}. This research can be briefly divided into two categories: the traditional analytical sampling-based method and the data-driven method.

\textbf{Traditional analytical sampling-based methods}\cite{ciocarlie2007dexterous,DBLP:conf/icra/GoldfederALP07,DBLP:conf/iros/HangSK14,DBLP:conf/icra/MillerKCA03,DBLP:conf/icra/PelossofMAJ04} sampled various grasp candidates and evaluated them based on certain metrics considering the physical properties of objects such as wrench space~\cite{DBLP:conf/icra/BorstFH04}. In general, both the object model and environment are assumed to be known in advance~\cite{DBLP:journals/ram/MillerA04}. Eigengrasp~\cite{ciocarlie2007dexterous} reduced the dimensions of grasp search space by performing principal component analysis (PCA) on grasping pose and configuration data. Although the reduction increases the efficiency of generating grasps, the search space of the random sampling process for grasps is still very huge. As a result, these sampling-based methods are less efficient in practical use.

\textbf{Data-driven methods} fall into one of two primary types.
The one is an extension of the traditional sampling-based method~\cite{DBLP:conf/iros/VarleyWWA15,DBLP:conf/icra/BorstFH04}. Instead of computing physical metrics, this method directly estimates grasp quality metrics from trained deep models. The grasp success rate can be greatly improved since traditional metrics cannot be computed accurately from an incomplete view of a novel object without any contact feedback. However, they are still dependent on known object models and exhibit the problem of huge sampling and search space.
The other data-driven method is performed in an end-to-end manner~\cite{DBLP:conf/iros/HangSK14,DBLP:conf/rss/LiuP0GM20,DBLP:journals/corr/abs-1908-04293,DBLP:conf/iros/LiuP0GM19,DBLP:conf/icra/KapplerBS15,DBLP:conf/iros/VarleyWWA15,mahler2017dex}. Specifically, this method takes the image or point cloud data of a grasped object as input and outputs a high-quality grasp. These approaches are able to effectively generate grasps and are robust to unknown objects. However, many can only handle a single object. Grasping may often fail due to the potential collision between the gripper and the environment.
Some recent work~\cite{DBLP:conf/icra/LiWL0LZ22,DBLP:conf/icra/LundellCLVWRMK21,DBLP:journals/corr/abs-2103-04783} predicts
collision-free \mbox{6-DoF} grasping in clutter using multi-finger grippers. They only classify the grasp types and do not take into account of the properties of multi-finger grasps. Our approach considers the gripper's physical structure and does not rely on the grasp types. Using a novel grasping representation and an end-to-end deep neural network based on contacts, our approach significantly reduces the search space for grasping and can generate reliable grasp poses.

\newcommand{\Rmatrix}{\mbox{${\bf{R}}$}\xspace}
\newcommand{\PC}{\mbox{${\bf{PC}}$}\xspace}

\section{Overview}\label{sec:overview}
\subsection{Problem Statement}
We address the problem of grasp generation for unknown objects in a cluttered environment with a multi-finger robotic hand. We take a single-shot of point cloud $\bf{PC}$ from a given viewpoint as input and predict a set of high-quality grasps $\bf{G}=\{\bf{p}, \bf{q}\}$ where $\bf{p}$ is hand pose and $\textbf{q}$ is hand joint configuration.

\begin{figure}[H]
\centering
\includegraphics[width=0.4\textwidth]{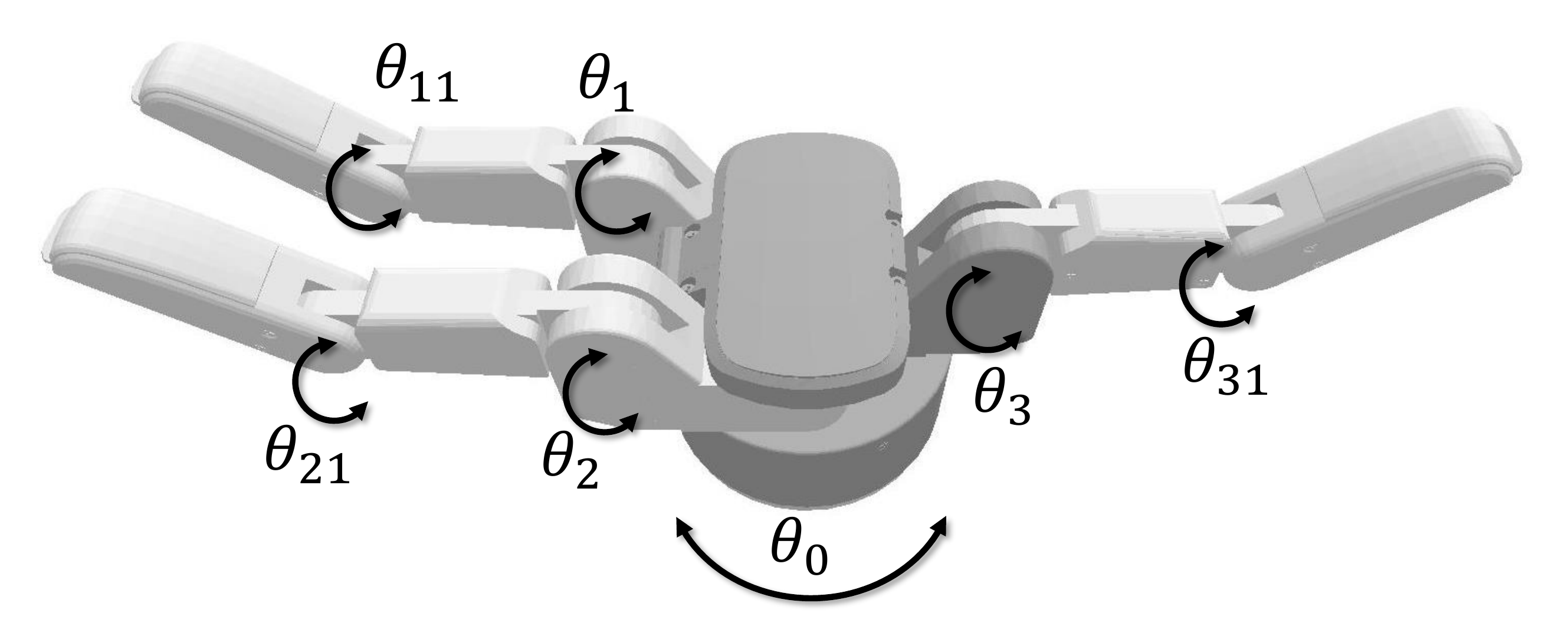}
\caption{A three-finger robotic hand.}
\label{fig:BarrettHand}
\end{figure}

In this paper, we use a \textit{BarrettHand} as our multi-finger dexterous hand, as shown in Fig.~\ref{fig:BarrettHand}.  The hand's pose consists of translation and orientation, denoted by
$
{\bf{p}}=\left\{x, y, z, \phi, \gamma, \varphi\right\}.
$
A hand joint configuration $\bf{q}$ consists of four inner joint angles, denoted by
$
{\bf{q}}=\left\{\theta_{0}, \theta_{1}, \theta_{2}, \theta_{3}\right\}.
$
Here, $\theta_{0}$ is the spread joint angle. $\theta_{1}$, $\theta_{2}$ and $\theta_{3}$ are the inner joint angles. The four outer joint angles $\theta_{11}$, $\theta_{21}$ and $\theta_{31}$ are dependent on their individual inner joints.

\subsection{Our Approach: Overview}

\begin{figure*}[thbp]
\centering
\includegraphics[width=0.9\textwidth]{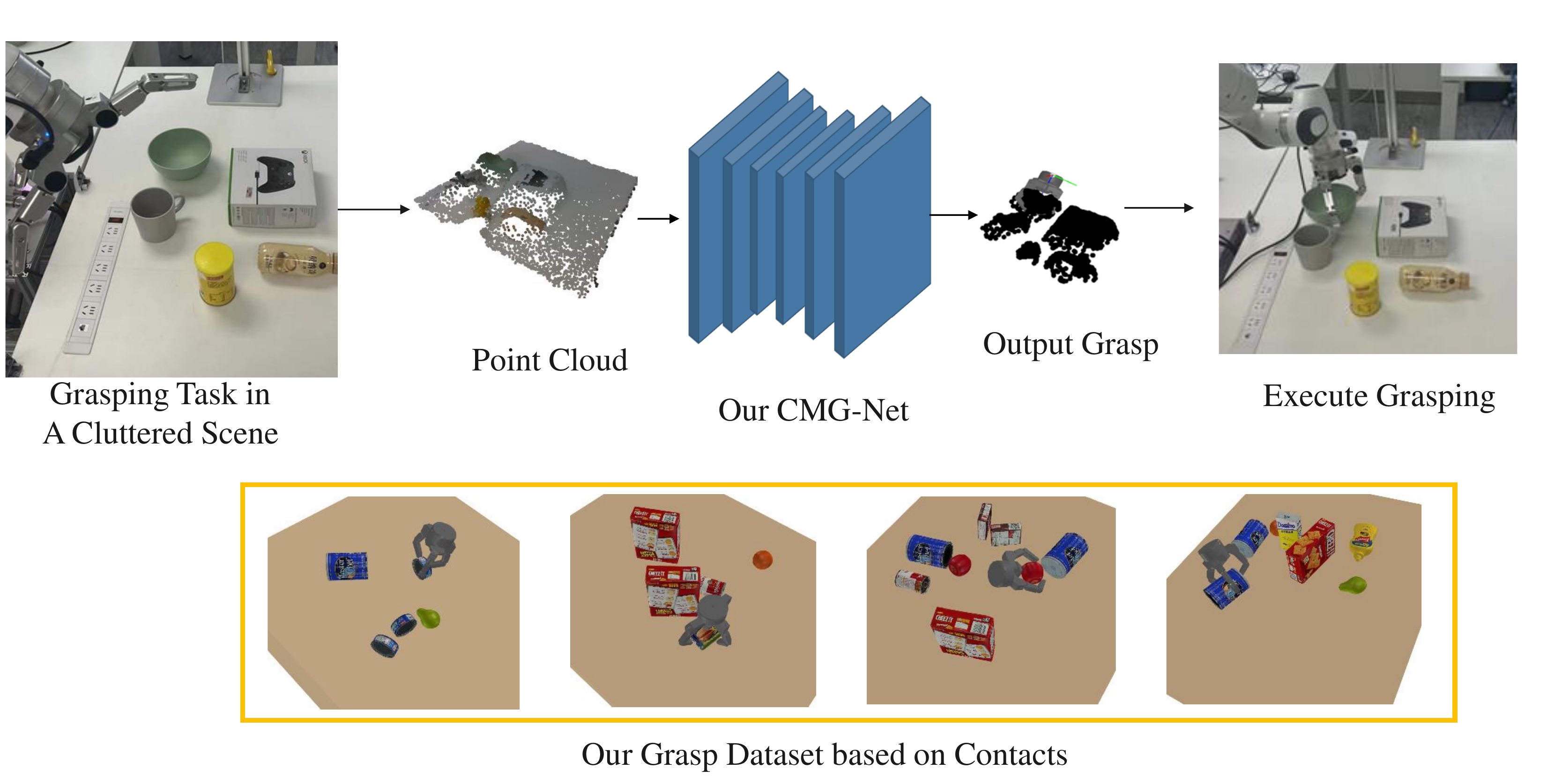}
\caption{The overview of our approach.}
\label{fig:overview}
\end{figure*}

Fig.~\ref{fig:overview} illustrates the overview of our approach. For a cluttered environment, we aim to grasp an unknown object using a multi-finger robotic hand. First, we obtain a point cloud captured from any viewpoint. Second, an end-to-end network, CMG-Net, is trained using our synthetic grasp dataset for the three-finger robotic hand based on contact representation. Third, we use the trained CMG-Net to obtain the final hand pose and grasp configuration. Finally, we execute the grasping task in real-world scenarios.

\section{Grasp Representation}\label{sec:representation}
Due to the ambiguous and discontinuous distribution of multi-finger grasps, it is difficult to directly regress a 6-DoF hand pose and a 4-DoF hand configuration in such a high dimensional space\cite{DBLP:conf/iccv/MousavianEF19,DBLP:conf/icra/LundellCLVWRMK21}.  Therefore, we define a novel grasp representation that can generate a constrained grasp space for our learning-based data-driven method.

\textbf{Multi-finger grasp representation}: For human hand grasp, Cutkosky~\cite{DBLP:journals/trob/Cutkosky89} suggested a taxonomy of two main categories: power and precision.
We observe that power grasp is not friendly to grasp objects on the desktop. In many scenarios, it requires the palm and fingers to entirely enclose an object, but this can easily cause collisions with the table during the hand closing action. In a real-world scenario, this may cause damage to expensive robotic hands.
Therefore, we consider the precision grasp when executing a grasping task.
This requires each finger to have contact with the object and but the contact between the palm, and the object is unnecessary.
For a dexterous hand like BarrettHand, whose two finger joints are coupled by a single motor, the contact points with the object for the precision grasp are usually at the end of fingertips.

\begin{figure}[h]
\centering
\includegraphics[width=0.45\textwidth]{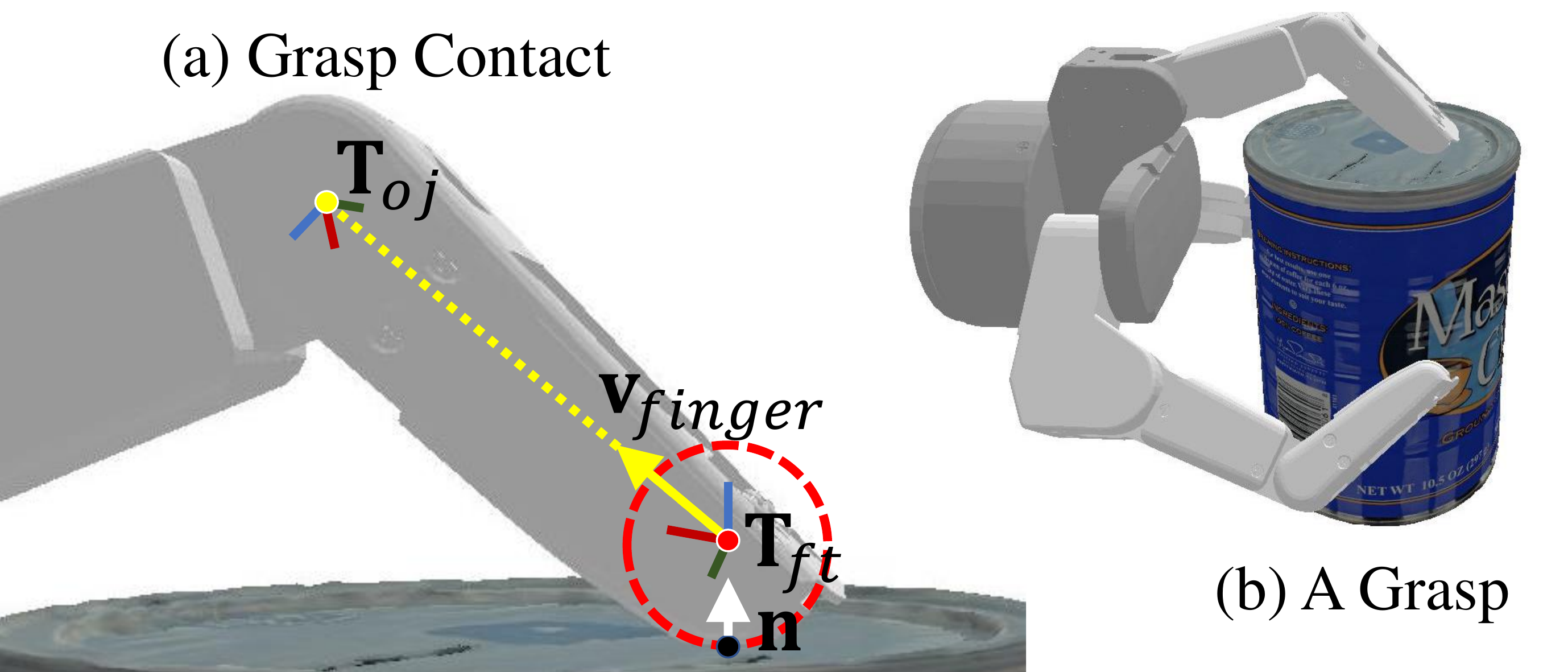}
\caption{Grasp representation based on contacts. (a) A contact between a fingertip and an object to be grasped. The black dot represents the contact point. The white arrow indicates the surface normal vector at the contact. The red circle highlights fingertip circle, the red point is the fingertip center and a coordinate frame ${\mathbf{T}}_{ft}$ is attached to the fingertip. The yellow arrow indicates the fingertip vector ${\mathbf{v}}_{finger}$. The yellow point is outer-joint position and a coordinate frame ${\mathbf{T}}_{oj}$ is attached. (b) A grasp example for a three-finger robotic hand.}
\label{fig:representation}
\end{figure}

To describe the contact between the fingertips and the object, we fit a fingertip end with a circle (Fig.~\ref{fig:representation}-(a)). The circle center at each fingertip (i.e., fingertip center) and its radius can be computed by the given gripper model. We denote this circle as the \textit{fingertip circle}. The local vector from the fingertip center to the origin of the outer joint is denoted as the fingertip vector ${\mathbf{v}}_{finger}$.
Based on this denotation, we generate a novel multi-finger grasping representation from the contact point to the corresponding finger joint pose to the hand pose and finally to the hand configuration.
We assume that the fingertip model of the dexterous hand in the side view can be matched with a particular circle.

Given a suitable contact point predicted on the object surface, the fingertip center ${\mathbf{t}}_{ft}$ in the world coordinate frame is computed as
\begin{align}
{\mathbf{t}}_{ft} = {\mathbf{t}}_{contact} + r\mathbf{n},
\end{align}
where ${\mathbf{t}}_{contact}$ is the position of the contact point, $\mathbf{n}$ is its surface normal vector and ${r}$ is the circle radius.
To compute the hand pose from the fingertip center, we attach a coordinate frame at the fingertip center.
At the contact point, The ${z}$-axis is the same as the normal vector.
To avoid confusion, we denote this vector as $\mathbf{v}=\left[v_1, v_2, v_3\right]^{T}$. Then the rotation matrix $\mathbf{R}_{ft}$ can be calculated by
\begin{align}
\mathbf{R}_{ft}=\left[\mathbf{R}^{1}, \left[0,-v_3, v_2\right]^{T},  \mathbf{v}\right],
\end{align}
where $\mathbf{R}^{1}=\left[0,-v_3, v_2\right]^T \times \mathbf{v}$. Then the hand pose $\mathbf{T}_{ft}$ is represented by the rotation matrix $\mathbf{R}_{ft}$ and the position of fingertip center $\mathbf{t}_{ft}$.

Then we compute a fixed-length vector from the fingertip center to the outer-joint position ($\mathbf{v}_{finger}$), by finger projections $\emph{x} \in \left[-1, 1\right]$ and $\emph{y} \in \left[-1, 1\right]$. $\emph{x}$ and $\emph{y}$ are the projections of the fingertip vector onto the $\emph{x}$ axis and $\emph{y}$ axis respectively in the fingertip center coordinate frame. The outer-joint position $\mathbf{t}_{oj}$ can be computed by
\begin{align}
\mathbf{t}_{oj}=\|\mathbf{v}_{finger}\| \left({\mathbf{R}}_{ft} \cdot \left[x, y, z\right]^T\right) + {\mathbf{t}}_{ft},
\end{align}
where $z=\sqrt{1-x^{2}-y^{2}}$.
Please refer to Fig.~\ref{fig:representation}-(a) for illustration. Here, $\emph{z}$ axis of the outer-joint coordinate frame is orthogonal to the fingertip vector $\mathbf{v}_{finger}$ and the $\emph{z}$ axis $\mathbf{v}_{z}$ of the fingertip center coordinate frame. Then we have the rotation matrix $\mathbf{R}_{oj}$ of the outer-joint coordinate frame
\begin{align}
\mathbf{R}_{oj} = \left[\mathbf{v}_{finger},
\mathbf{R}^2
, \mathbf{v}_{finger} \times \mathbf{v}_{z} \right] \cdot \mathbf{R}_{0},
\end{align}
where $\mathbf{R}^{2}=\mathbf{v}_{finger} \times \left[ \mathbf{v}_{finger} \times \mathbf{v}_{z} \right]$. $\mathbf{R}_{0}$ is a transformation matrix with a fixed angle of rotation around the $\emph{z}$ axis.

To further obtain the gripper pose from the outer-joint coordinate frame, we divided the four DoFs of the Barrett hand into two main joints($\theta_{ms}, \theta_{m}$) and two supporting joints($\theta_{s1}, \theta_{s2}$) according to whether they are involved in the calculation of the grasp pose or not. For example, if the contact as shown in Fig.~\ref{fig:representation}-(b) corresponds to finger $3$, the main joint $\theta_{m}$ is the joint $\theta_{3}$ and $\theta_{ms}$ is the spread joint $\theta_{0}$. Then the supporting joints $\theta_{s1}$ and $\theta_{s2}$ are $\theta_{1}$ for finger 1 and $\theta_{2}$ for finger 2, respectively. Therefore, the grasp pose is computed as
\begin{align} \label{xy_to_pose}
\mathbf{T}_{pose} = \mathbf{T}_{oj} \cdot \left(\mathbf{T}\right)^{-1},
\end{align}
where $\mathbf{T}_{oj}$ is composed by $\mathbf{R}_{oj}$ and $\mathbf{t}_{oj}$.
$\mathbf{T}$ is the transformation matrix from the base coordinate frame of the gripper to the out-joint coordinate frame. $\mathbf{T}$ can be computed from the main joints($\theta_{ms}, \theta_{m}$) using forward kinematics. Finally, we can combine the supporting joints to obtain the final grasp configuration. As a result, 10 dimensional grasp space is reduced to 6 dimensions, represented by
\begin{equation}
\mathbf{G}=\{\emph{x}, \emph{y}, \theta_{ms}, \theta_{m}, \theta_{s1}, \theta_{s2}\}.
\end{equation}

Using such a grasp representation, we can significantly reduce the grasp space to be searched, which greatly facilitates the learning process by analyzing the characteristics of the multi-finger hand structure. Object shape and hand structure are connected through contacts. This allow more reasonable predictions.

\section{Dataset Generation}\label{sec:dataset}
In this section, we explain the steps to generate our grasp dataset used in training.
For a given three-finger robotic hand, we first generate high-quality grasps for each object using \textit{GraspIt!}~\cite{DBLP:journals/ram/MillerA04} and keep those grasps that satisfy the grasp quality requirements. We obtain a grasp dataset consisting of contact information for each object.
Then we select a number of objects, randomly place them on the table, and examine all the grasps in a simulator. Those collision-free grasps and valid grasp poses remain. Finally, we place a camera in the scene and obtain a single-shot point cloud from a viewpoint. We map the grasp pose from the world coordinate system to the camera coordinate system.

\subsection{Single Object Grasp Dataset with Contacts}

To improve the diversity in shape, texture, and size, we select 80 objects from existing datasets~\cite{DBLP:conf/cvpr/FangWGL20,DBLP:journals/corr/CalliWSSAD15, DBLP:journals/corr/ChangFGHHLSSSSX15, xiangsapien}. Then we generate our grasp dataset for these objects in two steps.

First, we generate grasp poses and configurations for a single object. To uniformly sample the points on a single object for grasping, We down-sample the object mesh models to achieve a uniform distribution of sampling points
with their normal
in voxel space. For a sample, grasp candidates are searched in three dimensions $S_1 \times S_2 \times S_3$, where $S_1$ is the gripper depths, $S_2$ is the in-place rotation angle, and $S_3$ is the angle of spread joint. Given a set of $S_1 \times S_2 \times S_3$, we let fingers close in until they touch the object. We compute its $\epsilon$\emph{-quality}~\cite{ferrari1992planning} and save its grasp pose and configuration as a grasp annotation if the $\epsilon$\emph{-quality} is greater than a specified threshold. In our dataset, we generate 15000 high-quality grasp annotations for each object.

Second, we generate the contacts between the robotic hand and the objects for those high-quality grasps. Here, we ignore a grasp if its fingertips have no contact with the given object. We use the $k$-means algorithm to compute the clustering of the contact points and thus generate the corresponding contact point for each fingertip.

\subsection{Scene Grasp Dataset Generation}
To generate grasps in a scene from an arbitrary viewpoint, we first take a number of objects randomly from the object dataset and place them on the table in a stable pose. Second, for each object in the scene, we retrieve grasps from the grasp dataset and perform grasping tasks in the simulation environment. We filter out those grasps exhibiting collisions and invalid poses. Third, we place a camera to capture the scene from any viewpoint. The captured point cloud is used to obtain the grasp poses $\mathbf{G}$ and contact positions in the camera coordinate system.

\section{Grasp Network: CMG-Net}
\label{sec:CMG}
\begin{figure}[ht]
\centering
\includegraphics[width=0.98\columnwidth]{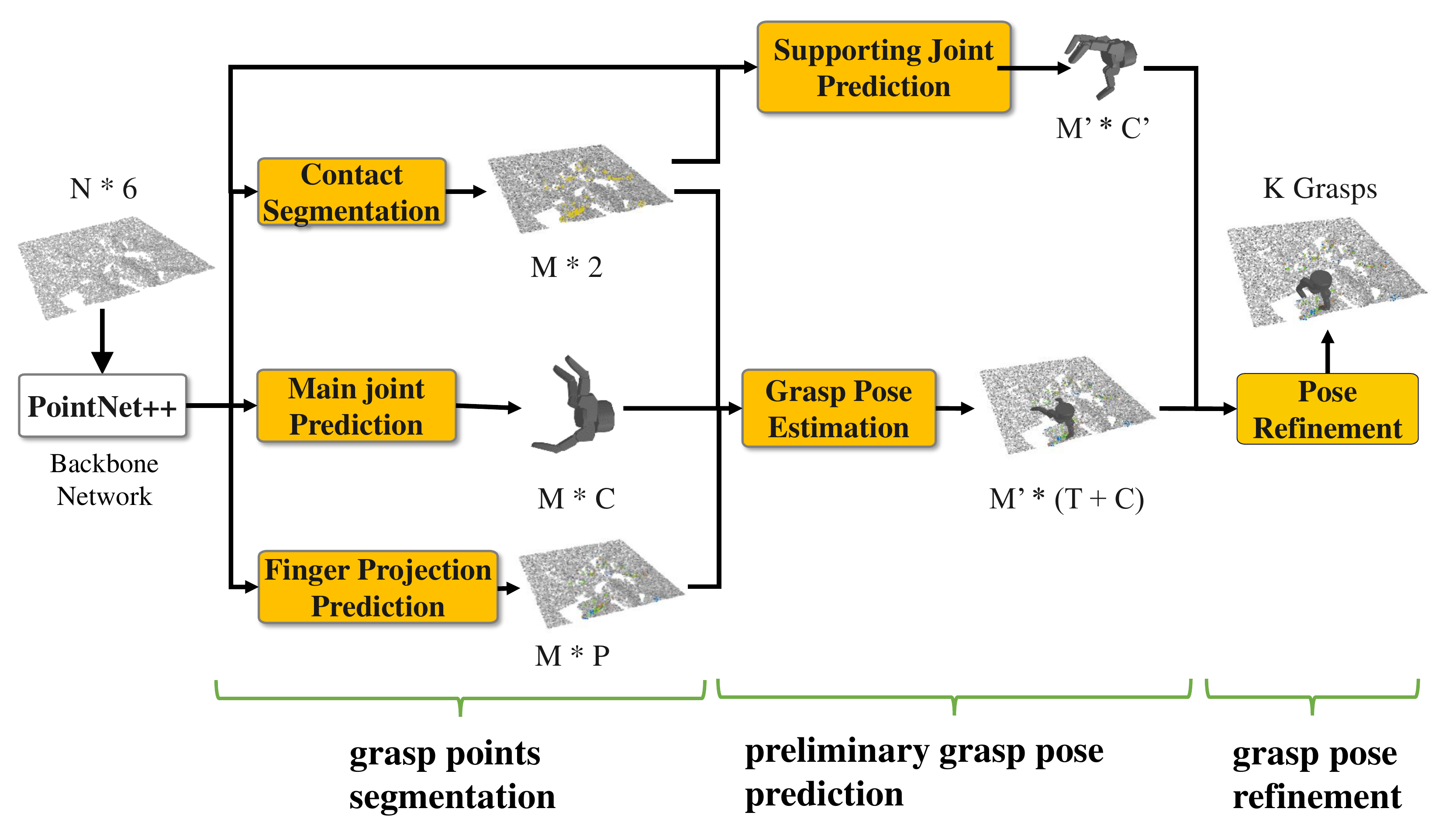}
\caption{The CMG-Net framework.}
\label{fig:CMGNet}
\end{figure}

In this section, we introduce a simple grasp pose estimation network: CMG-Net (Contact-based Multi-finger Grasping Network).
For an input point cloud, we aim to generate a set of grasp poses.
Fig.~\ref{fig:CMGNet} shows CMG-Net's structure, consisting of three stages,
1) grasp points segmentation, which extracts features from the point cloud by PointNet++~\cite{DBLP:conf/nips/QiYSG17}, and then classifies the contact points to determine whether they are graspable. And we predict the finger ID (e.g., finger 1, 2, 3) and finger projection corresponding to the possible contact points.
2) preliminary grasp pose prediction, which predicts an initial pose using our proposed grasp representation,
and 3) grasp pose refinement, which refines the grasping pose by predicting the supporting joints.

\subsection{Learning to Predict Grasp Points}
\label{sub:LearningPredictGrasp}

We use PointNet++ \cite{DBLP:conf/nips/QiYSG17} as our backbone due to its simplicity and high success rate on various challenging tasks.
We use its set-abstraction (SA) layers and feature propagation (FP) layers to build an asymmetric U-shaped network with an encoder and a decoder.
The encoder uses multiple SA blocks to extract abstract features from point clouds and the decoder utilizes an equal number of FP blocks to gradually interpolate the abstracted features.

We also incorporate the point normals into the network. The actual input data size is increased to $N \times 6$, where $N$ is the number of points.
The extracted point cloud features, as shown in Fig.~\ref{fig:CMGNet}, are used by the \emph{contact segmentation module} to predict whether a point is graspable.
We design a cross-entropy loss $\mathcal{F}_{cls}$ for classification training:
\begin{align}
\mathcal{L}_{gp}=\mathcal{F}_{cls}\left(c^{p}, {c^{g}}\right),
\end{align}
where $c^{p}$ is the prediction result of segmentation head for each grasp sample and $c^{g}$ is its corresponding label.

\subsection{Grasp Pose Estimation}
We further predict the corresponding joint angles of the given dexterous hand based on the graspable points and combine them with finger projection to calculate a preliminary grasp pose using Eq.~\ref{xy_to_pose}.
Specifically,
We use \emph{Finger Projection Prediction}, which contains a multi-layer perceptron (MLPs) network with fully connected layers, ReLU, and batch normalization, to predict finger projection $x^p$ and $y^p$. The finger projection $x$ and $y$ will be predicted as a group $xy^p$. The loss function is
\begin{equation}
\mathcal{L}_{fp}=\frac{1}{N_{\text{c}}} \sum_i{\left\|{xy}_i^p-{xy}_i^{l}\right\|} \delta_{\text{c}},
\end{equation}
where $N_\mathrm{c}$ is the total number of contact points. $xy_i^{l}$ is the ground truth finger projection from the contact points, $\delta_c$ is an indicator, and it yields 1 if $xy_i$ corresponds to contact points and 0 otherwise.
The joint angle $\theta_{m}$ is evenly divided into $n_m$ bins. In our implementation, we set the subdivision angle $\phi_{m} = \frac{7\pi}{9}$.
For each grasp sample, we calculate its bin classification label ${bin}_{m}^l$ and residual label  ${res}_{m}^l$ as follows
\begin{equation}
\begin{aligned}
{bin}_{m}^l &=
\left\lfloor\frac{{\theta}_m}{\phi_m}\right\rfloor, \\
{res}_m^l &=\frac{1}{{\phi}_m}\left({\theta}_m^l-\left({bin}_m  {\phi}_m+\frac{{\phi}_m}{2}\right)\right).
\end{aligned}
\end{equation}

The loss of main joint is formulated as
\begin{equation} \label{bin_loss}
\mathcal{L}_{m}=\mathcal{F}_{c l s}\left(bin_m^{p}, {bin_m^{l}}\right) + \mathcal{F}_{res}\left(res_m^{p}, {res_m^{l}}\right),
\end{equation}
where $bin_m^l$ and $res_m^l$ are ground-truth bin assignment and residual for a given grasp $p$. $bin_m^p$ and $res_m^p$ are their corresponding predicted values.

Analogically, the spread joints $\theta_{ms}$ has the same form
{\small
\begin{align}
\mathcal{L}_{ms}=\mathcal{F}_{c l s}\left(bin_{ms}^{p}, {bin_{ms}^{l}}\right) + \mathcal{F}_{res}\left(res_{ms}^{p}, {res_{ms}^{l}}\right).
\end{align}
}

According to Eq.~\ref{xy_to_pose}, we finally obtain a preliminary grasp pose $\mathbf{G}=\{\emph{x}, \emph{y}, \theta_{m},\theta_{ms}\}$.

\subsection{Grasp Pose Refinement}
To improve grasp quality and create realistic grasp poses, a joint prediction layer is used to guarantee the accuracy of supporting joints $\theta_{s1}$ and $\theta_{s2}$.
More specifically, $\theta_{s1}$ and $\theta_{s2}$ are uniformly divided into $n_{s1}$ and $n_{s2}$ bins, respectively.
For each point $p_i$, we compute the bin classification label and the residual label as
\begin{align}
\mathcal{L}_{s_i}=\mathcal{F}_{c l s}\left(bin_{s_i}^{p}, {bin_{s_i}^{l}}\right) + \mathcal{F}_{res}\left(res_{s_i}^{p}, {res_{s_i}^{l}}\right),
\end{align}
where $i\in \{1, 2\}$ represents one supporting joint. $bin_{s_i}$ and $res_{s_i}$ are classification and residual labels, respectively.

\subsection{Total Loss}
The total loss function for training consists of three major components: grasp point prediction $\mathcal{L}_{gp}$, preliminary grasp generation using finger projection $\mathcal{L}_{fp}$, and grasp refinement using joint prediction $\mathcal{L}_{joint}$, which can be formally expressed as
\begin{align}
\mathcal{L}_{total}=\alpha\mathcal{L}_{gp} + \beta\mathcal{L}_{fp} + \gamma\mathcal{L}_{joint},
\end{align}
where
\begin{align}
\mathcal{L}_{joint} = \gamma_1\mathcal{L}_{m} + \gamma_2\mathcal{L}_{ms} + \gamma_3\mathcal{L}_{s_i}.
\end{align}

In our implementation, we set $\alpha=1$, $\beta=\gamma=5$, $\gamma_1=\gamma_2=\gamma_3=1$, and use the cross-entropy loss~\cite{shannon1948mathematical} and Huber (smooth-L1) loss~\cite{DBLP:journals/pami/RenHG017} for all classification and regression tasks, respectively.

\section{Experiments and Results}\label{sec:results}
In this section, we first describe the experimental setup, including the dataset usage, evaluation metric, and implementation details. Then we demonstrate our experimental results in simulation and real scenarios. In our experiments and comparison, we observe that our approach can generate dense and robust grasps with high success and completion rates compared to the baseline approaches.

\subsection{Dataset Usage}
Our dataset has more than 80 categories distributed among 5000 different scenarios.
The network outputs a point-wise grasp pose and hand configurations for a single viewpoint cloud. We consider a grasp attempt is performed successful if the object can lift at least 30$cm$ high.

\subsection{Simulation Experiments}
We implement our CMG-Net using PyTorch \cite{DBLP:conf/nips/PaszkeGMLBCKLGA19} on NVIDIA  GPUs. Our end-to-end network is optimized using the Adam optimizer with a batch size of 32 and a learning rate of 0.004. Our input is a point cloud converted from the depth map captured using a depth camera. We randomly down-sample the point cloud to retain a reasonable number of points (e.g., 20,000).

\begin{table}[h]
\centering
\caption{Comparison in Simulation}
\begin{tabular}{c|cccc}
             & SR(\%) & CR(\%) & Quality \\ \midrule
GraspIt!     &   48 &   51 &     0.63    \\ \cmidrule(lr){1-4}
Multi-FinGan &  56  &   64 &    0.76    \\ \cmidrule(lr){1-4}
Ours         &  76  &  81  &   0.86
\end{tabular}\label{main_results}
\end{table}

\subsubsection{Evaluation Metrics}
To demonstrate the performance, we introduce three metrics~\cite{DBLP:conf/icra/LundellCLVWRMK21,DBLP:journals/corr/abs-2103-04783,qin2020s4g}:
		 \emph{Grasp Success Rate (SR)} is the rate of the number of successful grasps to the number of total attempts.
		 \emph{Grasp Completion Rate (CR)} is the rate of the number of objects grasped to the total number of objects after 1.5 times the number of grasp attempts.
		 \emph{Grasps Quality} often refers to $\epsilon$\emph{-quality}  metric~\cite{ferrari1992planning},  representing the radius of the largest 6D ball centered on the origin that can be surrounded by the convex hull of the wrench space~\cite{borst2004grasp}.

\subsubsection{Results}
\begin{figure}[htbp]
  \subfloat[\label{sim_small}For small object.]{
      \includegraphics[width=0.45\columnwidth]{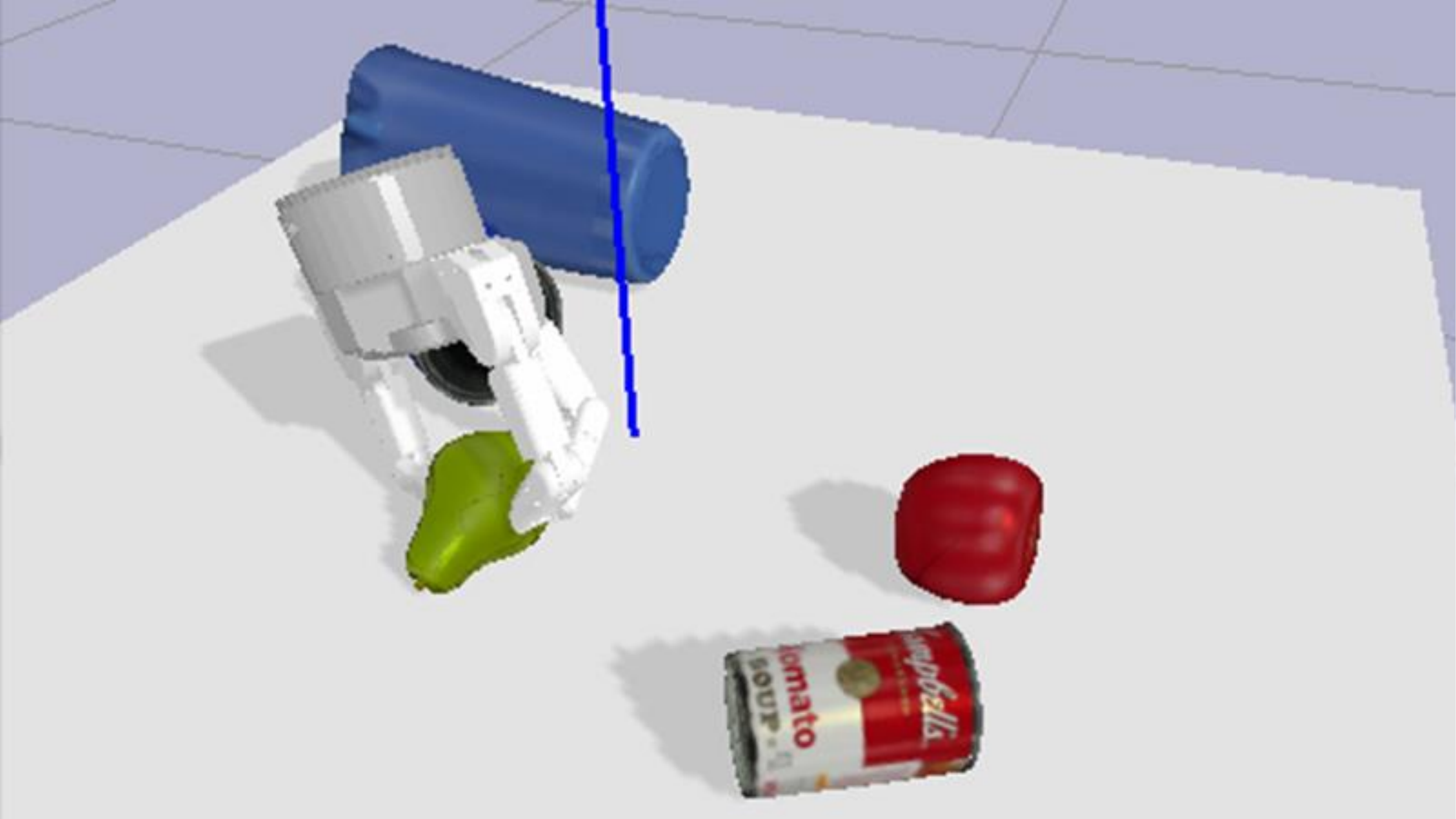}
  }
  \subfloat[\label{sim_com}In cluttered scenes.]{
      \includegraphics[width=0.45\columnwidth]{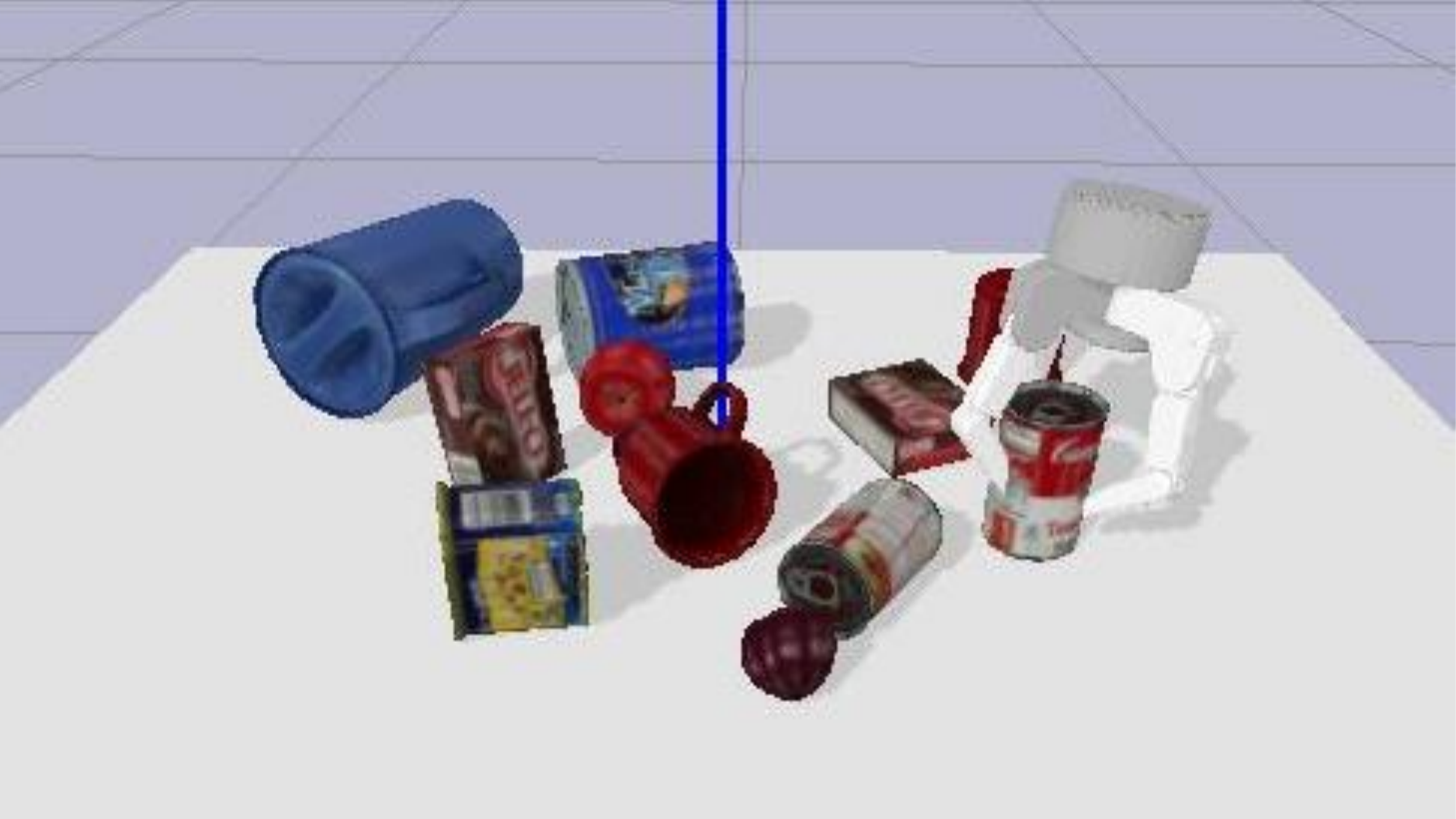}
  }
  \caption{Grasp poses in different situations.}
  \label{sim_res}
\end{figure}

The experimental results (Table~\ref{main_results}) show that our network can be well incorporated with the proposed representation. In comparison, our approach shows significant improvement against $GraspIt!$ and Multi-FinGan in terms of success rate, completion rate, and grasp quality.
Fig.~\ref{sim_res} shows a few simulation experiment results, demonstrating our method can handle small objects very well and generate high-quality grasps for complex scenes.

\subsubsection{Ablation Studies}
To improve the quality of handling small objects, we combine the SA and FP layers in PointNet++ through a U-shape network for feature extraction. In addition, we reduce the required regression dimensions by predicting the finger projection, which significantly improves the prediction quality of grasps and reduces the training cost. Table~\ref{ase} shows the enhancements resulting from each of the two modules. From the results, we can see that our proposed finger projection can achieve significant improvement.
\begin{table}[htbp]
\centering
\caption{Ablation Study}
\begin{tabular}{l|cccc}
              & U-Shaped & Finger-Pro & SR(\%) & Quality \\ \midrule
Baseline      &                  &              &  39      &    0.56  \\ \cmidrule(lr){1-5}
+U-Shaped     &    \checkmark    &              &  46      &    0.65  \\ \cmidrule(lr){1-5}
+Finger-Pro   &                  & \checkmark   &  72      &    0.80  \\ \cmidrule(lr){1-5}
Ours          &    \checkmark    & \checkmark   &  76      &    0.86
\end{tabular}\label{ase}
\end{table}

\subsection{Real-world Experiments}

\begin{figure}[ht]
\centering
\includegraphics[width=0.98\columnwidth]{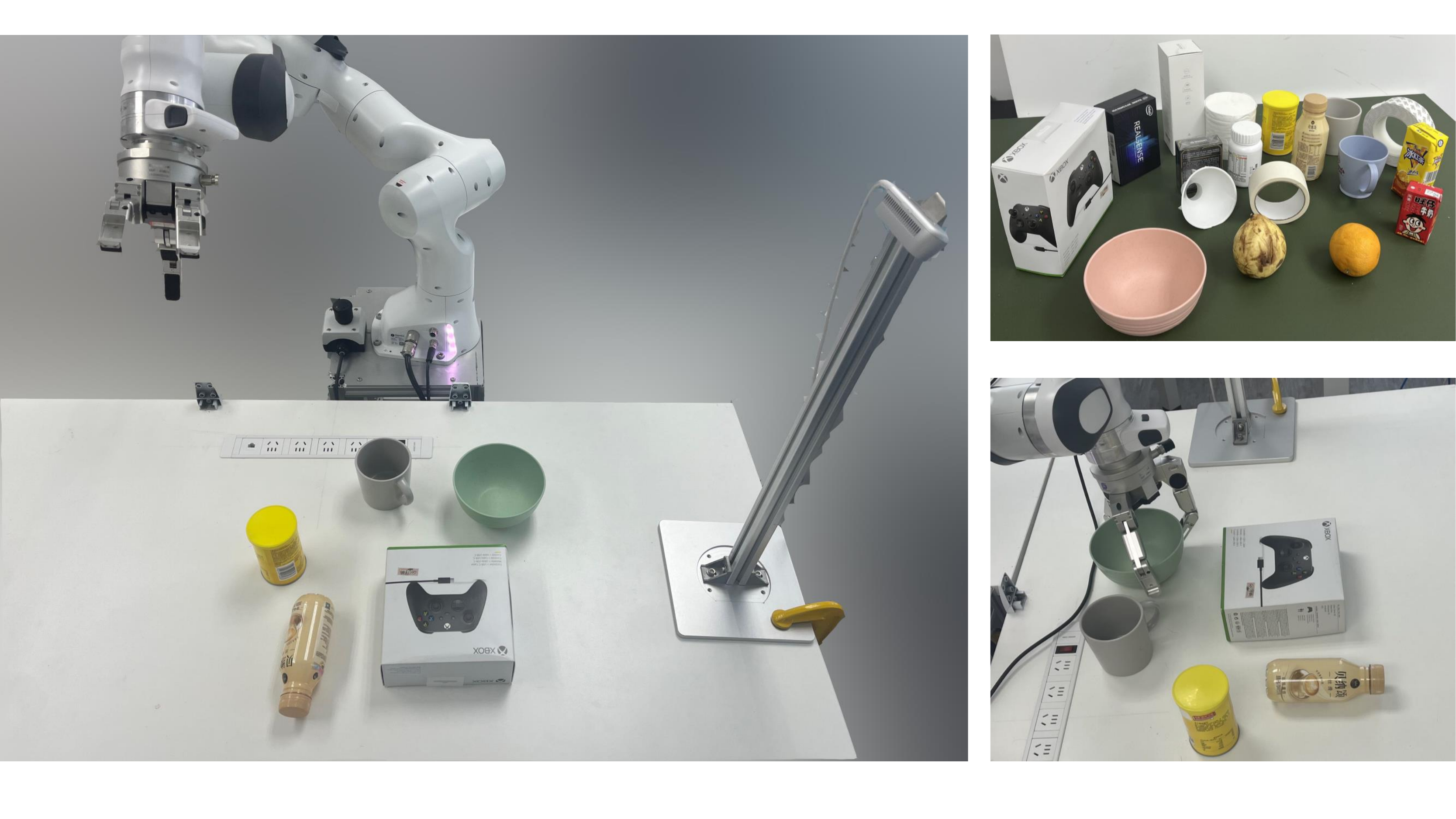}
\caption{A real-world experimental environment for object manipulation using a three-finger robotic hand and the given objects.}\label{real_e}
\end{figure}

We demonstrate that the network trained from the simulation data works well in a real environment. We use a \textit{Franka Emika Panda} equipped with a three-finger dexterous hand in our real-world experiment. We use an \textit{Intel RealSense D435} camera to capture the depth images. In order to remove the background and desk points for grasping prediction,
we perform depth image segmentation using a segmentation network~\cite{DBLP:conf/corl/XiangXMF20}.
Then the object point clouds are fed into our CMG-Net to obtain the final grasp. For each input, we keep the top 20 best grasps tried in the real field scene. A grasp is successful when an object can be picked up and stably moved to a specified location. Fig.~\ref{real_e} shows a snapshot of the real-world experiment environment and the objects grasped during the experiment.
There are three experimental scenarios with 3, 6, and 9 objects to be grasped, respectively. We obtain the average of 6 trials for each scenario. In these real scenarios, the objects are all novel for robotic hand.

The experimental result is shown in Table~\ref{real_world_results}. We obtained a success rate up to $74.4\%$ and a completion rate up to $86.1\%$ in the 6-object scenario. The results look similar even for the 9-object scenario. Our work demonstrates a good adaptability for a cluttered environment with many unknown objects.

\begin{table}[h]
\centering
\caption{Real Hardware Experiment Results}
\begin{tabular}{c|ccc}
Objects      & 3 & 6 & 9 \\ \midrule
SR(\%)       &  76.5  & 74.4  &  72.0    \\ \cmidrule(lr){1-4}
CR(\%)       &  88.9  & 86.1  &  83.3
\end{tabular}\label{real_world_results}
\end{table}

\section{Conclusion}\label{sec:conc}
We present a novel contact-based grasp representation for a three-finger robotic dexterous hand and propose an end-to-end neural network (CMG-Net) to predict the grasp pose for unknown objects from only one single-shot point cloud in a cluttered environment. To train CMG-Net, we generate a large-scale synthetic dataset for three-finger grasps. We have compared CMG-Net against the state-of-the-art methods and observed 20\% performance improvements in success rate.

There are a few limitations in our work. We mainly consider the grasping for static and rigid objects. This may limit its applications, especially in household scenes.
Our approach requires the fingertips having contacts with the given object. However, in some grasps, this requirement may be not necessary.
In the future, we would like to investigate the grasping representation and prediction for dynamical and soft objects.
though it is much more difficult to formulate the problem.
We would like to extend our approach to handle arbitrary grasps (i.e., with or without fingertip contacts).

Moreover, other possible future work may incorporate the grasping network into robotic manipulation skills, achieving more complex tasks with multi-finger hand, e.g., assembling or house cleaning.
\section*{Acknowledgments}
This work is supported by the National Key R\&D Program of China under Grant 2021ZD0114501, the Science and Technology Innovation Action Plan of Shanghai under Grant 22511105400, and partially supported by Ascend AI Computing Platform and CANN (Compute Architecture for Neural Networks).

\bibliographystyle{IEEEtran}
\bibliography{reference.bib}

\end{document}